\documentclass[sigconf]{acmart}

\AtBeginDocument{%
  \providecommand\BibTeX{{%
    \normalfont B\kern-0.5em{\scshape i\kern-0.25em b}\kern-0.8em\TeX}}}

\copyrightyear{2022}
\acmYear{2022}
\setcopyright{acmlicensed}
\acmConference[WWW '22 Companion] {Companion Proceedings of the Web Conference 2022}{April 25--29, 2022}{Virtual Event, Lyon, France.}
\acmBooktitle{Companion Proceedings of the Web Conference 2022 (WWW '22 Companion), April 25--29, 2022, Virtual Event, Lyon, France}
\acmPrice{15.00}
\acmISBN{978-1-4503-9130-6/22/04} 
\acmDOI{10.1145/3487553.3524194}

\begin{document}

\title{Geometric and Topological Inference for Deep Representations of Complex Networks}

\author{Baihan Lin}
\email{baihan.lin@columbia.edu}
\affiliation{%
  \institution{Columbia University}
  \city{New York}
  \state{NY}
  \country{USA}
  \postcode{10027}
}

\renewcommand{\shortauthors}{Lin}

\begin{abstract}
Understanding the deep representations of complex networks is an important step of building interpretable and trustworthy machine learning applications in the age of internet. Global surrogate models that approximate the predictions of a black box model (e.g. an artificial or biological neural net) are usually used to provide valuable theoretical insights for the model interpretability. In order to evaluate how well a surrogate model can account for the representation in another model, we need to develop inference methods for model comparison. Previous studies have compared models and brains in terms of their representational geometries (characterized by the matrix of distances between representations of the input patterns in a model layer or cortical area). In this study, we propose to explore these summary statistical descriptions of representations in models and brains as part of a broader class of statistics that emphasize the \textit{topology} as well as the \textit{geometry} of representations. The topological summary statistics build on topological data analysis (TDA) and other graph-based methods. We evaluate these statistics in terms of the sensitivity and specificity that they afford when used for model selection, with the goal to relate different neural network models to each other and to make inferences about the computational mechanism that might best account for a black box representation. These new methods enable brain and computer scientists to visualize the dynamic representational transformations learned by brains and models, and to perform model-comparative statistical inference. 
\end{abstract}




\maketitle

\section{Introduction}

In this work, we propose a series of statistical inference techniques to compare complex networks in terms of summary-statistical descriptions that capture their topological and geometric structure. 
Consider a set of \textit{N} individuals in a social network. Their relationships can be characterized by their pairwise degrees of separation in the network, by dissimilarities of their profiles, or by geographical distances between their homes. Each of these topological and geometric summaries can be expressed in a square matrix of pairwise relationships. The same summary statistics can be used to characterize representations of \textit{N} visual images in a layer of a neural network or in a cortical area of human brain, indicating to what extent different images are dissimilar in multivariate neural representational space (geometry) and which stimuli share neighbor relationships (topology) in that space. Based on persistent homology, prior work in topological data analysis has enabled expressive graph visualizations. Here we focus on geometric and topological descriptors for formal inferential comparisons between multiple representations or networks. These methods enable us to infer, for example, to what extent geographical proximity and profile similarity are related to the degrees of separation in a social network, and whether proximity is significantly more related to the degrees of separation than profile similarity. Similarly, these methods support assessing to what extent the representation of photos in a neural network is similar to the representation in another neural network or a brain region, and whether one of the networks provides a better model of the brain representation.  

To integrate geometric (distance-based) and topological (graph-based) characteristics and embed both in a consistent inferential framework, we introduce a nonlinear monotonic transformation of distances, which provides a family of statistics combining geometric and topological information. Our applications include neuroscience \cite{lin2019visualizing}, collaborative filtering and independence testing \cite{lin2018adgtic}.
 
Other than graph-based network properties, we also investigate higher-order topological inference and information-theoretical measures. For instance, similar to the evolving nature of social networks and preferential graphs, in single-cell genomics, the lack of a formal link between cell-cell cohabitation and its emergent dynamics into cliques during development has hampered our understanding of how cell populations evolve across time. We introduce higher-order simplicial structure as a new summary statistic, and discover that these networks contain an abundance of cliques of single-cell profiles bound into cavities that guide the emergence of more complicated habitation forms \cite{simpArch,sctsa}. To further disentangle information flow, we develop a mathematical filtration technique to compute nerve balls in a dual metric of space and time \cite{ttda} and a information-theoretical measure among network modules \cite{lin2019neural,lin2022regularity}.

\section{Problem Setting and significance}

This work aims to solve the following two analytical problems.

\textbf{Aim 1: Develop novel analyses of representational \textit{geometry} and \textit{topology}.} Visualizations of neural dynamics can reveal representational transformations and, thus, neural computations. Classical studies have used linear dimensionality reduction methods (e.g., PCA \cite{pearson1901liii}, jPCA \cite{churchland2012neural}, dPCA \cite{kobak2016demixed}, tensor components analysis \cite{Williams2018}, ``hypothesis-guided'' dimensionality reduction (HDR) \cite{lara2018a}) to embed high-dimensional trajectories of neural population activity in visualizations. However, linear dimensionality reduction may distort the geometry, collapsing dimensions that contribute to the discriminability of representational patterns. We develop extensions of multidimensional scaling (MDS) \cite{buja2008data} and TDA \cite{carlsson2009topology} for visualization of network dynamics that are optimized to accurately reflect the representational geometry and topology. These methods provide an important complement to existing techniques by forgoing the simplicity of a linear subspace in exchange for a better reflection of the degree to which information about stimuli and task-relevant latent variables is represented for readout by downstream modules.

\textbf{Aim 2: Investigate the representational \textit{dynamics} and \textit{information} of recurrent networks.} Recurrent neural nets (RNNs), as the generalized form of neural nets, are not only useful for learning the latent structure of sequences, but also provide a candidate computational model for how the human and animal brains perform a variety of tasks. As an RNN engages a dynamic world, information about the recent history of inputs is accumulated, dynamically compressed, and propagated along the cyclic computational graph. Classical studies in dynamical modeling recently supported a direct mathematical link of recurrent networks with other popular deep learning architectures such as residual network (ResNet) and convolutional networks (ConvNet) \cite{liao2016bridging}. Investigating the representational dynamics of RNNs in information flow and accumulation within and across layers is critical to deep learning theory and to the development of better architectures that provide appropriate inductive biases to enable a faster learning and faster dynamic convergence during inference. We explore a broad set of summary statistics (e.g. SVCCA \cite{raghu2017svcca}, CKA \cite{kornblith2019similarity} and the novel geo-topological summary statistics from Aim 1) and visualization methods. 

Different summary statistics are then evaluated by an objective benchmark that reflects the degree to which they enable accurate inferences about data-generating model, when data are sparsely sampled
and models are trained from different random seeds than the data-generating one. The goal is to select the best descriptors to capture underlying invariant aspects of representational geometry and topology along the temporal dimension of RNN dynamics (across unfolded time steps, layers, learning stages and task states).


\textbf{Impact will be threefold}: (1) Understanding the representational signatures of RNNs help us compare and improve architectures, which enable engineering advances. (2) Visualizations reveal the latent task-related states, contributing to explainable AI by providing a more abstract functional description of the computations performed by RNNs. (3) These new methods enable us to test RNNs as models of dynamic computations in brains. This addresses a major challenge in computational neuroscience: how to connect increasingly rich and detailed measurements of brain activity to increasingly complex and powerful neural net models?


\section{Methods}

We introduce a range of new analysis and visualization methods.
\vspace{-1em}

\subsection{Develop novel analyses of representational \textit{geometry} and \textit{topology}}



The representational geometry is defined by the representational dissimilarity matrix (RDM) as the summary statistic or, equivalently, by the second moment of the activity profiles. The RDM constitutes an intermediate level summary statistic: It discards some of the information contained in the distribution of activity profiles, but retains more information than a linear decoding approach. Here we explore discarding additional information and considering the representational \textit{topology} alongside the representational \textit{geometry}. 

There are two motivations for exploring this, one theoretical and one data-analytical. 
From a theoretical perspective, 
the local geometry determines which input stimuli the representation renders indiscriminable, which it discriminates, but places together in a cluster, and which it places in different neighborhoods. The global geometry of the clusters (whether two stimuli are far or very far from each other in the representational space) may be less relevant to computation: In a high-dimensional space a set of randomly placed clusters will tend to afford linear separation of arbitrary dichotomies among the clusters \cite{rigotti2013,kushnir2018} independent of the exact global geometry. Like a storage room, a representational space may need to co-localize related things, while the global location of different categories may be arbitrary.
From a data-analytical perspective, conversely, small distances may be unreliable given the various noise sources that may affect the measurements. From both theoretical and data-analytical perspectives, it seems possible that focusing our sensitivity on a particular range of distances turns out to be advantageous because it reduces the influence of noise and/or arbitrary inter-individual variability (e.g., of the global geometry) that does not reflect computational function.

\begin{figure}[tb]
\centering
\includegraphics[width=1.0\linewidth]{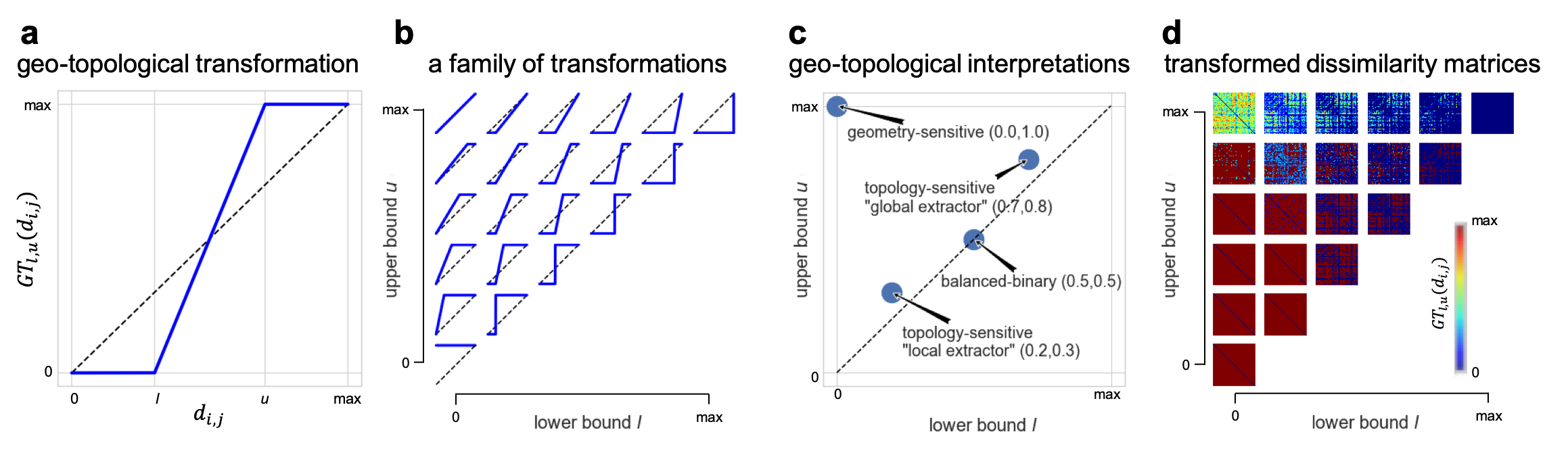}
\vspace{-1em}
\caption{\footnotesize \textbf{Representational Geo-Topology Analysis (RGTA)}. (a) nonlinear monotonic transform of distances; (b) a family of transforms; (b) interpretations of learned transformations; (c) effect on the distance matrices.
}\label{fig:RGTA}
\vspace{-1em}
\end{figure}

We introduce a novel approach called representational geo-topology analysis (RGTA), which builds on the literature on topological methods (e.g., persistent homology \cite{zomorodian2005computing}, TDA mapping \cite{carlsson2009topology}).
The utility of a focus on topology derives from the idea of persistence, where unreliable features are discarded to reveal the underlying structure in the data. In order to summarize the idiosyncrasies of individual brain mechanisms and highlight the representational properties that are key to their computational function, we would like to find an optimal upper distance threshold $u$ above which we consider stimuli distinct, but do not consider differences between distances meaningful (i.e., the stimuli are disconnected in the graph capturing the topology at the appropriate scale). In order to suppress noise, we would like to find a lower threshold $l $ below which we consider stimuli as co-localized (i.e., the stimuli collapse into the same node in the graph). Between two thresholds we place a continuous transition to retain some geometrical sensitivity in the range where it matters (Figure~\ref{fig:RGTA}).

\textbf{\textit{Optimizing representational geo-topological analysis for RNNs.}} When applying 
RGTA to a multi-layer neural network, we will optimize thresholds $l$ and $u$, so as to best select the data-generating model representation. Model selection is based on a comparison of the RGTA summary statistics estimated from data generated by some model to those from separately trained instances of the same and different network architectures (different random seeds, different training samples). This kind of simulation is critical, because the fact that ground truth (i.e. the data-generating model) is known enables us to search our family of geo-topological summary statistics for those that provide the greatest power to adjudicate among models.
One possibility is that the ideal setting is $l =0$, $u=max$, i.e., the original RDM. In that case the result would be that there is no benefit to geo-topological transformation of the RDM in the context of the network modules (e.g. model layers) and the experimental conditions studied. However, if region identification benefits from the geo-topological transform, then we will explore it further for visualization and inferential analyses. To the degree that the greater invariance of topological summary statistics helps match corresponding representations despite idiosyncratic parameters of individual models, inferential comparisons of deep network models with brains should utilize these summary statistics, rather than the traditional statistics
of
representational geometry.

\textbf{\textit{Higher-order topological simplicial analysis with spatio-temporal filtration and witness bootstrapping.}} 
``Traditional'' TDA applications usually focus on the low-dimensional graph visualization and the persistent homology of the data (i.e. computing the Betti numbers or barcodes up to dimension 2), because interpreting the physical meaning of the geometry and higher dimensional persistent modules is a conceptual challenge. We propose to use instead the simplicial architecture of the networks. Simplicial analysis was first introduced to study human brain connectomes \citep{reimann2017cliques}, where each connected pairs of neurons are considered an edge to create a graph and the numbers of Rips-Vietoris simplices in dimensions up to 7 can be computed. However, the filtration challenge of deriving a graph from the distance-based data by choosing the best threshold, hinders the practical application of such simplicial analysis in these point cloud data. In \cite{ttda}, we tackle this issue by introducing a bootstrapping procedure from the point cloud and impose an additional constraint in the temporal dimension. These developments we can efficiently estimate the inter-module interaction between computational units within our topics of interest.


\vspace{-1em}

\subsection{Investigate the representational \textit{dynamics} and \textit{information} of recurrent networks}


Neural computations unfold over time. Visualizations of network dynamics can help us understand how representations are transformed and gain insight into the computational process. Static visualizations of individual time frames can be animated by applying them in a time-windowed style to render neural dynamics as a movie. However, it is also important to visualize neural dynamics in static images. Classical studies have used dimensionality reduction methods (notably PCA and supervised variations including jPCA \cite{churchland2012neural}, dPCA \cite{kobak2016demixed}, and HDR \cite{lara2018a}) to embed high-dimensional trajectories of neural population activity in 2d and 3d visualizations. This is a powerful approach. However, for rich sets of representational states (e.g. many visual stimuli or distinct movements), linear dimensionality reduction may collapse neural states that are untangled in the high-dimensional representation \cite{russo2018}. 
This motivates alternative visualization methods that aim to represent the pairwise discriminabilities of the representational states (e.g. of sensory stimuli). MDS offers one such method, which optimizes the accuracy with which pairwise distances in the high-dimensional space are reflected in the visualization.
We develop extensions of MDS for visualizing dynamic representational geometries. The analyses of representational dynamics will use the summary statistics of geometry and topology developed in Aim 1. We can relax the requirement of MDS for models to accurately reflect the geometry (e.g. pairwise distances) and require only that models accurately reflect the topology (e.g. neighborhoods) of representational space. 

\begin{figure}[tb]
\centering
\includegraphics[width=0.9\linewidth]{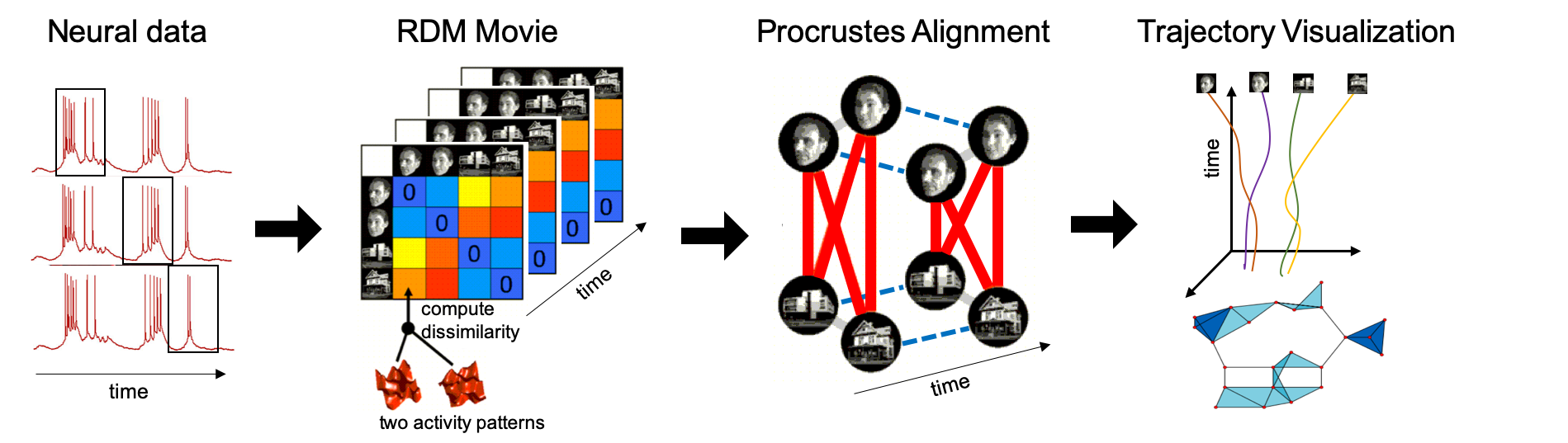}
\caption{\footnotesize \textbf{Pipeline of 2d+time representational dynamic analysis.} 
}\label{fig:RepDynPipeline}
 \vspace{-1em}
\end{figure}

\textbf{\textit{Ride along with a dynamic representational geometry.}} Einstein famously imagined riding alongside a beam of light. This thought experiment, a mental visualization, rendered stationary the most rapid dynamic process, and was a major inspiration that contributed to a scientific breakthrough. To visualize neuronal population dynamics, we may similarly choose either a static or a moving reference frame. 
Neural population dynamics \cite{mante2013context,churchland2012neural,hung2005fast} 
are often visualized using dimensionality reduction methods such as PCA and its extensions \cite{mante2013context,churchland2012neural,lara2018conservation,kobak2016demixed} to find a single 2d or 3d linear subspace in which to view the trajectories. This approach of linear-subspace trajectories is well motivated by its simplicity and interpretability. Certain distinctions may be collapsed, but an advantage is that the approach equally represents \textit{within}-time-point and \textit{between}-time-point relationships among representational patterns. This enables us to visualize neural population response for even a single experimental condition as a trajectory. 

Here we introduce an alternative approach, in which we travel along with the ensemble of high-dimensional points that correspond to different experimental conditions (e.g., different stimuli or different movements performed by the animal). Rather than embedding all neural activity patterns across all time points in a single space, we first seek to represent the relationships between the patterns separately at each time point. We then align the embeddings across time points so as to \textit{minimize} motion of the ensemble as a whole and render more apparent the changes of the representational geometry or topology, i.e. which stimuli become distinct and which collapse and how their global distances develop. 
Any visualization method including linear subspace methods and the methods from Aim 1 could be used to obtain a separate configuration for each time frame. To illustrate the concept of a dynamic visualization with a moving frame of reference, let us consider MDS. MDS is a natural choice for visualizing geometries because it minimizes the distortion of the distances in the visualization (using the metric stress cost function here). The MDS cost function explicitly penalizes the collapsing of points that are distant in the the original space, e.g., neural patterns that represent distinct states \cite{russo2018}.

\begin{figure}[tb]
\centering
\includegraphics[width=0.95\linewidth]{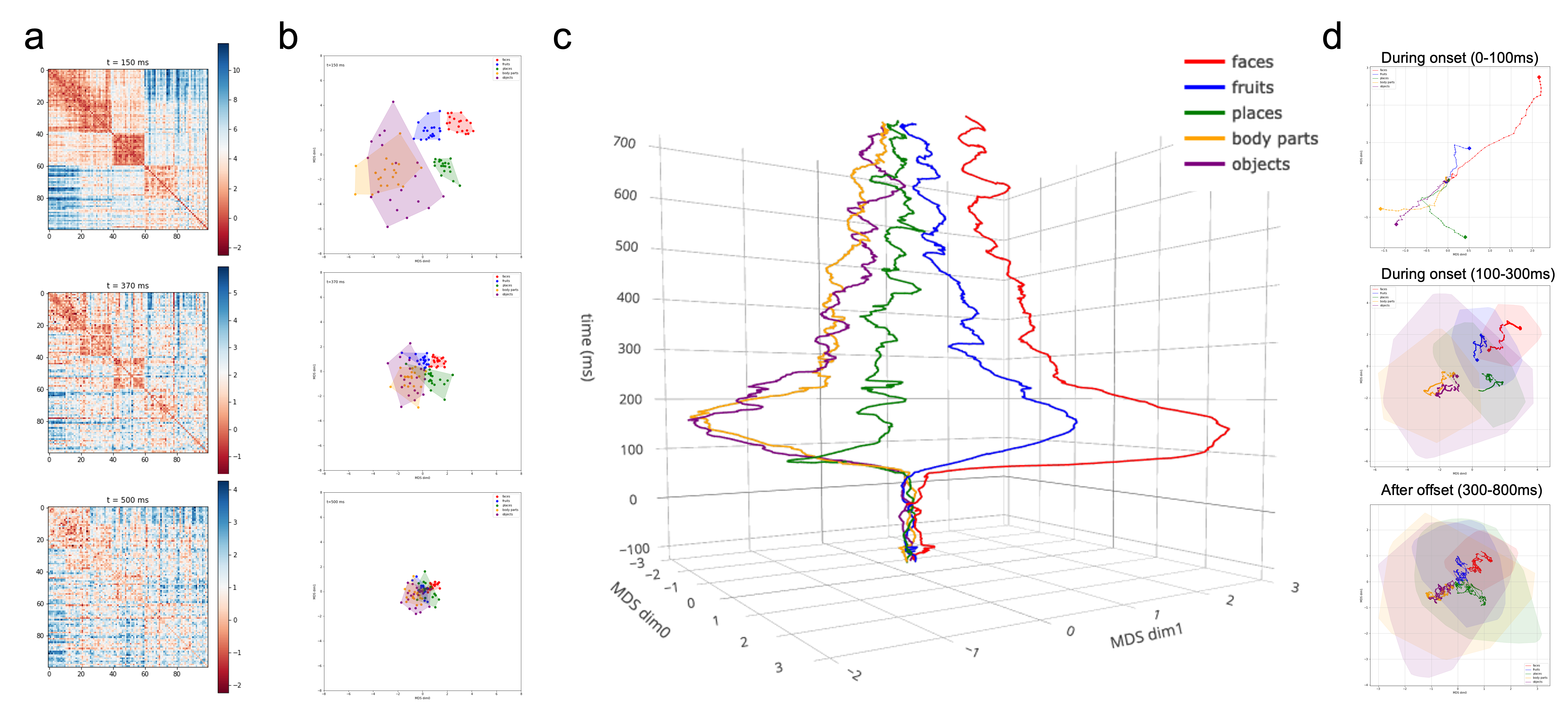}
\caption{\footnotesize MDS alignment. (a, b) RDM and MDS at example time points (onset at 0ms, during onset at 150ms, and after offset at 370ms); (c) \textbf{3d trajectory} of Procrustes-aligned MDS over time; (d) dynamic synopsis of 3 major stages.}\label{fig:3d}
 \vspace{-1em}
\end{figure}

Figure \ref{fig:RepDynPipeline} shows the sequence of steps of the proposed procedure. First each time frame of neural activity is analyzed separately. This could involve, for example, spike counts performed in a temporal sliding window to obtain a neural population activity pattern for each experimental condition. Second, for each time frame, the representational dissimilarity is estimated \cite{kriegeskorte2019a} for each pair of experimental conditions and assembled in the representational dissimilarity matrix (RDM). Stacking these RDMs along the time dimension results in an RDM movie. Third, each RDM is separately subjected to MDS to obtain, for each time frame, a 2d or 3d configuration of the points for visualization. Fourth, the configurations are aligned to each other using Generalized Procrustes Analysis (GPA), and finally, the aligned configurations are presented together. The alignment implements the moving frame of reference, which minimizes the motion of the ensemble as a whole in the visualization from one time point to the next. We can present the low-dimensional embedding of the dynamics as a movie (using visualization time to represent neural activity time) or as a static figure. For a static figure, we must choose how to indicate the time frame. In Figures \ref{fig:RepDynPipeline} and \ref{fig:3d}, we designate one dimension of the visualization space as representing time and used the remaining two orthogonal dimensions for the relationships among stimuli within a time frame.

\textbf{\textit{Assessing model similarity of RNNs in different tasks.}} Following a similar experimental setup as \cite{maheswaranathan2019universality}, we train populations of thousands of networks, with different variants of RNN architectures such as RNN, Long-Short-Term-Memory (LSTM) \cite{hochreiter1997long}, Gated Recurrent Unit (GRU) \cite{cho2014learning} and bidirectional versions of the above \cite{schuster1997bidirectional}, and characterize their nonlinear dynamics. We consider a range of neuroscientifically and computationally motivated tasks: signal-transform tasks such as frequency-cued sine wave and K-bit flip-flop \cite{sussillo2013opening}, computer vision tasks such as object classification and location, and logic tasks such as time-windowed logic gates (e.g. XOR). To select the best statistics for the model similarity, we compare and evaluate several metrics: (a) representational-space-based metrics (of dimension $n_{cond} \times n_{channel}$) such as SVCCA, CKA and projection weighted CCA (PWCCA) \cite{morcos2018insights}, and (b) RDM-based metrics \cite{kriegeskorte2013representational} (of dimension  $n_{cond} \times n_{cond}$) such as cosine distance, Euclidean distance, and correlation distance with Pearson's r, Spearman's $\rho$, and Kendall's $\tau$. The model pairs being considered are different RNN networks with different architectures, different RNN layers, the same layer but at different time steps, RNN networks in different training stages, as well as RNNs trained in one task and tested for generalization to a different task.

\begin{figure}[tb]
\centering
\includegraphics[width=0.95\linewidth]{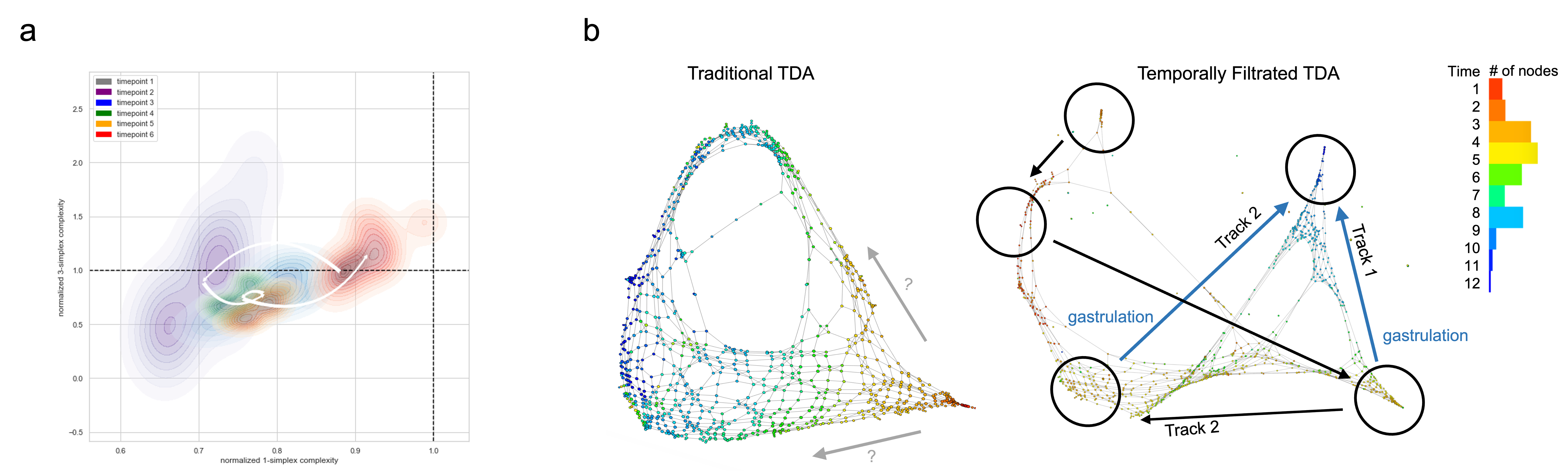}
\caption{\footnotesize (a) Higher-order simplicial statistics offer insights in biological time-series data; (b) temporal filtration disentangles the information flow.
 }\label{fig:ttda}
  \vspace{-1.5em}
\end{figure}

\begin{figure}[tb]
\centering
\includegraphics[width=0.95\linewidth]{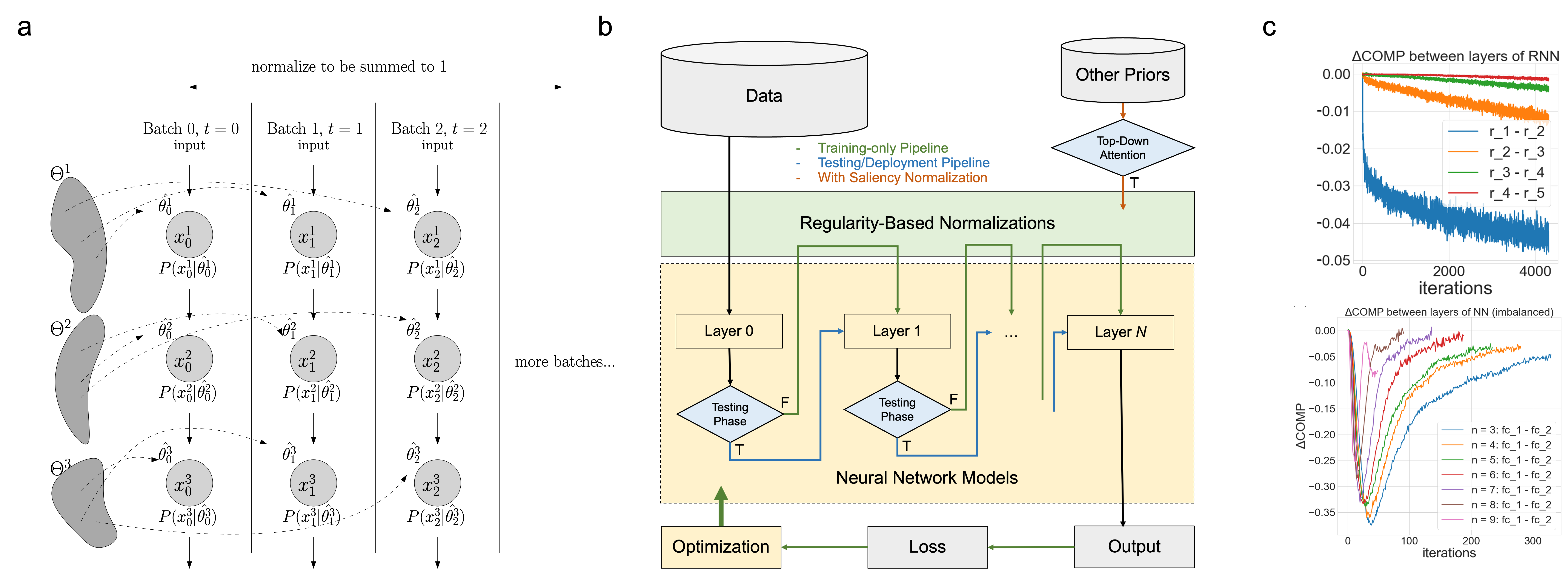}
\caption{\footnotesize (a) Neural network as a model selection process; (b) pipeline to compute MDL; (c) example information probing of neural nets.
 }\label{fig:rn}
  \vspace{-1.5em}
\end{figure}

\textbf{\textit{Assessing model-brain similarity, based on neural population activity data.}} The methods will be applied to a wide range of neural population activity data sets, acquired in different species (mice, marmoset monkeys, macaque monkeys) during a variety of tasks (visual perception, odor discrimination, motor tasks). We pursue both data-driven visualizations of neural population dynamics and model-driven analyses, where the aim is to adjudicate among alternative RNN models as models of the dynamics and computational function of biological neural networks. 

\section{Results}

We present the results from four aforementioned method directions:

\textbf{\textit{Adaptive geo-topological independence criterion.}} Testing two potentially multivariate variables for statistical dependence on the basis finite samples is a fundamental statistical challenge. In \cite{lin2018adgtic}, we explore a family of tests that adapt to the complexity of the relationship between the variables, promising robust power across scenarios. Building on the distance correlation, we introduce a family of adaptive independence criteria based on nonlinear monotonic transformations of distances and show that these criteria, like the distance correlation and RKHS-based criteria, provide dependence indicators. Given this effective measure, we propose a class of adaptive (multi-threshold) test statistics, which form the basis for permutation tests. These tests empirically outperform some of the established tests in average and worst-case statistical sensitivity across a range of univariate and multivariate relationships.

\textbf{\textit{Procrustes-aligned multidimensional scaling.}} In \cite{lin2019visualizing}, we demonstrate the feasibility of riding along with a dynamic representational geometry, by analzying a biologcal neural net from a macaque recording dataset.
Figure~\ref{fig:3d} shows the aligned 2d MDS configurations as a function of time (3rd axis). Neural population activity was averaged for each category. The dynamics of the category-centroid relationships are clearly observable. The category centroids emanate from the origin, becoming clearly separate around 80 ms after stimulus onset and reach a maximum distinction at around 150 ms. While the stimulus is still on (duration: 300 ms), the trajectories gradually converge over time in an oscillatory fashion after the stimulus offset (at 300ms). One important next step will be to compare these trajectories with those of recurrent networks. 

\textbf{\textit{Topological simplicial analysis with temporal filtration.}} In \cite{sctsa,simpArch}, we propose a topological simplicial analysis (TSA) pipeline as an exploratory inquiry to solve these two challenges: (1) with the algebraic geometry's definitions of forming higher-order simplices, we can potentially interpret that cliques of higher orders indicates operational units of higher order; (2) with the bootstrapping techniques to sample from the data points collected at each sub-level, we can scale the analysis to large single cell datasets and compare groups of cells quantitatively. In addition, in \cite{ttda} we heuristically define a technique to extract the temporal skeleton of the developmental processes, called temporally filtrated TDA, and show that the developmental trajectories of cells can be better revealed in this approach comparing to traditional TDA mapping. 

\textbf{\textit{Information flow within network modules with minimum description length.}} To put this direction in perspective, we mention that this line of work is a series of three. The entry point is \cite{lin2019neural}, where we propose a perspective to understand neural network optimization as a partially observable model selection problem. In our subsequent work \cite{lin2022regularity}, we introduce the details of how to approximate the minimum description length (MDL) between neural network layers and demonstrate that using MDL as the regularity information is useful, from an engineering angle, for neural networks to learn from certain input data distributions. By comparing with other theoretical tools such as mutual information, we envision it as a tool to understand how information propagated between network modules, a venue not widely explored previously, as in \cite{lin2022regularity,neco_mdl}. In our third stage,
we aim to bridge the information-theoretical tool (e.g. MDL) with the group actions. We have this intuition that traditional usage of mutual information mostly deals with statistical association, while MDL quantifies the efficiency of the transformation function in the communication system, i.e. the degree to which the groups are symmetric. If we can formulate the theoretical grounds for this intuition, we might be able to show that over the time of the optimization process, the transformation function (as a manifold) converges to an invariant geometric structure, reflected by a higher and higher communication efficiency. 

\vspace{-1em}
\section{Conclusion}

In summary, we propose a series of statistical inference techniques to compare complex networks in terms of summary-statistical descriptions that capture their topological and geometric structure, in order to characterize the temporal evolution of dynamic networks and information flow of computational graphs. There are four main research outcomes: (1) an adaptive independence test with geo-topological transforms; (2) a topological simplicial framework for high-throughput time-series data; (3) the representational similarity analysis for robust model inference of artificial and biological neural nets; (4) a MDL estimate to quantify information flow among network modules. Future work include applying these techniques on other complex networks (e.g. social networks and recommendation systems) and providing physical interpretations for the geometric and topological objects emerged from these complex networks.

\begin{acks}
I thank Dr. Nikolaus Kriegeskorte, who supervised and assisted the research, and the reviewers for their comments to the manuscript. 
\end{acks}
\vspace{-0.25em}
\bibliographystyle{ACM-Reference-Format}
\bibliography{main}


\begin{thebibliography}{34}


\ifx \showCODEN    \undefined \def \showCODEN     #1{\unskip}     \fi
\ifx \showDOI      \undefined \def \showDOI       #1{#1}\fi
\ifx \showISBNx    \undefined \def \showISBNx     #1{\unskip}     \fi
\ifx \showISBNxiii \undefined \def \showISBNxiii  #1{\unskip}     \fi
\ifx \showISSN     \undefined \def \showISSN      #1{\unskip}     \fi
\ifx \showLCCN     \undefined \def \showLCCN      #1{\unskip}     \fi
\ifx \shownote     \undefined \def \shownote      #1{#1}          \fi
\ifx \showarticletitle \undefined \def \showarticletitle #1{#1}   \fi
\ifx \showURL      \undefined \def \showURL       {\relax}        \fi
\providecommand\bibfield[2]{#2}
\providecommand\bibinfo[2]{#2}
\providecommand\natexlab[1]{#1}
\providecommand\showeprint[2][]{arXiv:#2}

\bibitem[Buja~et. al.(2008)]%
        {buja2008data}
\bibfield{author}{\bibinfo{person}{Andreas Buja~et. al.}}
  \bibinfo{year}{2008}\natexlab{}.
\newblock \showarticletitle{{Data visualization with multidimensional
  scaling}}.
\newblock \bibinfo{journal}{\emph{J Comput Graph Stat}} \bibinfo{volume}{17},
  \bibinfo{number}{2} (\bibinfo{year}{2008}), \bibinfo{pages}{444--472}.
\newblock


\bibitem[Carlsson(2009)]%
        {carlsson2009topology}
\bibfield{author}{\bibinfo{person}{Gunnar Carlsson}.}
  \bibinfo{year}{2009}\natexlab{}.
\newblock \showarticletitle{{Topology and data}}.
\newblock \bibinfo{journal}{\emph{Bull. A. M. S.}} \bibinfo{volume}{46},
  \bibinfo{number}{2} (\bibinfo{year}{2009}), \bibinfo{pages}{255--308}.
\newblock


\bibitem[Cho~et. al.(2014)]%
        {cho2014learning}
\bibfield{author}{\bibinfo{person}{Kyunghyun Cho~et. al.}}
  \bibinfo{year}{2014}\natexlab{}.
\newblock \showarticletitle{Learning phrase representations using RNN
  encoder-decoder for statistical machine translation}.
\newblock \bibinfo{journal}{\emph{arXiv preprint arXiv:1406.1078}}
  (\bibinfo{year}{2014}).
\newblock


\bibitem[Churchland~et. al.(2012)]%
        {churchland2012neural}
\bibfield{author}{\bibinfo{person}{Mark~M Churchland~et. al.}}
  \bibinfo{year}{2012}\natexlab{}.
\newblock \showarticletitle{{Neural population dynamics during reaching}}.
\newblock \bibinfo{journal}{\emph{Nature}} \bibinfo{volume}{487},
  \bibinfo{number}{7405} (\bibinfo{year}{2012}), \bibinfo{pages}{51}.
\newblock


\bibitem[Hochreiter and Schmidhuber(1997)]%
        {hochreiter1997long}
\bibfield{author}{\bibinfo{person}{Sepp Hochreiter} {and}
  \bibinfo{person}{J{\"u}rgen Schmidhuber}.} \bibinfo{year}{1997}\natexlab{}.
\newblock \showarticletitle{Long short-term memory}.
\newblock \bibinfo{journal}{\emph{Neural computation}} \bibinfo{volume}{9},
  \bibinfo{number}{8} (\bibinfo{year}{1997}), \bibinfo{pages}{1735--1780}.
\newblock


\bibitem[Hung~et. al.(2005)]%
        {hung2005fast}
\bibfield{author}{\bibinfo{person}{Chou~P Hung~et. al.}}
  \bibinfo{year}{2005}\natexlab{}.
\newblock \showarticletitle{{Fast readout of object identity from macaque
  inferior temporal cortex}}.
\newblock \bibinfo{journal}{\emph{Science}} \bibinfo{volume}{310},
  \bibinfo{number}{5749} (\bibinfo{year}{2005}), \bibinfo{pages}{863--866}.
\newblock


\bibitem[Kobak~et. al.(2016)]%
        {kobak2016demixed}
\bibfield{author}{\bibinfo{person}{Dmitry Kobak~et. al.}}
  \bibinfo{year}{2016}\natexlab{}.
\newblock \showarticletitle{{Demixed principal component analysis of neural
  population data}}.
\newblock \bibinfo{journal}{\emph{Elife}}  \bibinfo{volume}{5}
  (\bibinfo{year}{2016}), \bibinfo{pages}{e10989}.
\newblock


\bibitem[Kornblith~et. al.(2019)]%
        {kornblith2019similarity}
\bibfield{author}{\bibinfo{person}{Simon Kornblith~et. al.}}
  \bibinfo{year}{2019}\natexlab{}.
\newblock \showarticletitle{Similarity of neural network representations
  revisited}.
\newblock \bibinfo{journal}{\emph{arXiv preprint arXiv:1905.00414}}
  (\bibinfo{year}{2019}).
\newblock


\bibitem[Kriegeskorte and Diedrichsen(2019)]%
        {kriegeskorte2019a}
\bibfield{author}{\bibinfo{person}{Nikolaus Kriegeskorte} {and}
  \bibinfo{person}{J{\"{o}}rn Diedrichsen}.} \bibinfo{year}{2019}\natexlab{}.
\newblock \showarticletitle{{Peeling the Onion of Brain Representations}}.
\newblock \bibinfo{journal}{\emph{Annual Review of Neuroscience}}
  \bibinfo{volume}{42}, \bibinfo{number}{1} (\bibinfo{year}{2019}),
  \bibinfo{pages}{407--432}.
\newblock
\showISSN{0147-006X}


\bibitem[Kriegeskorte and Kievit(2013)]%
        {kriegeskorte2013representational}
\bibfield{author}{\bibinfo{person}{Nikolaus Kriegeskorte} {and}
  \bibinfo{person}{Rogier~A Kievit}.} \bibinfo{year}{2013}\natexlab{}.
\newblock \showarticletitle{{Representational geometry: integrating cognition,
  computation, and the brain}}.
\newblock \bibinfo{journal}{\emph{Trends Cogn. Sci.}} \bibinfo{volume}{17},
  \bibinfo{number}{8} (\bibinfo{year}{2013}).
\newblock


\bibitem[Kushnir and Fusi(2018)]%
        {kushnir2018}
\bibfield{author}{\bibinfo{person}{Lyudmila Kushnir} {and}
  \bibinfo{person}{Stefano Fusi}.} \bibinfo{year}{2018}\natexlab{}.
\newblock \showarticletitle{{Neural Classifiers with Limited Connectivity and
  Recurrent Readouts}}.
\newblock \bibinfo{journal}{\emph{J. Neurosci.}} \bibinfo{volume}{38},
  \bibinfo{number}{46} (\bibinfo{year}{2018}), \bibinfo{pages}{9900--9924}.
\newblock
\showISSN{0270-6474}


\bibitem[Lara et~al\mbox{.}(2018)]%
        {lara2018a}
\bibfield{author}{\bibinfo{person}{Antonio~H Lara}, \bibinfo{person}{John~P
  Cunningham}, {and} \bibinfo{person}{Mark~M Churchland}.}
  \bibinfo{year}{2018}\natexlab{}.
\newblock \showarticletitle{{Different population dynamics in the supplementary
  motor area and motor cortex during reaching}}.
\newblock \bibinfo{journal}{\emph{Nature Communications}} \bibinfo{volume}{9},
  \bibinfo{number}{1} (\bibinfo{year}{2018}), \bibinfo{pages}{2754}.
\newblock
\showISSN{2041-1723}


\bibitem[Lara~et. al.(2018)]%
        {lara2018conservation}
\bibfield{author}{\bibinfo{person}{Antonio~H Lara~et. al.}}
  \bibinfo{year}{2018}\natexlab{}.
\newblock \showarticletitle{{Conservation of preparatory neural events in
  monkey motor cortex regardless of how movement is initiated}}.
\newblock \bibinfo{journal}{\emph{Elife}}  \bibinfo{volume}{7}
  (\bibinfo{year}{2018}).
\newblock


\bibitem[Liao and Poggio(2016)]%
        {liao2016bridging}
\bibfield{author}{\bibinfo{person}{Qianli Liao} {and} \bibinfo{person}{Tomaso
  Poggio}.} \bibinfo{year}{2016}\natexlab{}.
\newblock \showarticletitle{{Bridging the gaps between residual learning,
  recurrent neural networks and visual cortex}}.
\newblock \bibinfo{journal}{\emph{Preprint arXiv:1604.03640}}
  (\bibinfo{year}{2016}).
\newblock


\bibitem[Lin(2019a)]%
        {simpArch}
\bibfield{author}{\bibinfo{person}{Baihan Lin}.}
  \bibinfo{year}{2019}\natexlab{a}.
\newblock \showarticletitle{Cliques of single-cell {RNA}-seq profiles reveal
  insights into cell ecology during development and differentiation}. In
  \bibinfo{booktitle}{\emph{ISMB}}. \bibinfo{address}{Basel, Switzerland}.
\newblock


\bibitem[Lin(2019b)]%
        {lin2019neural}
\bibfield{author}{\bibinfo{person}{Baihan Lin}.}
  \bibinfo{year}{2019}\natexlab{b}.
\newblock \showarticletitle{Neural networks as model selection with incremental
  MDL normalization}. In \bibinfo{booktitle}{\emph{Human Brain and Artificial
  Intelligence}}. Springer, \bibinfo{pages}{195--208}.
\newblock


\bibitem[Lin(2022a)]%
        {neco_mdl}
\bibfield{author}{\bibinfo{person}{Baihan Lin}.}
  \bibinfo{year}{2022}\natexlab{a}.
\newblock \showarticletitle{A note on estimating MDL in neural networks}.
\newblock \bibinfo{journal}{\emph{under review}} (\bibinfo{year}{2022}).
\newblock


\bibitem[Lin(2022b)]%
        {lin2022regularity}
\bibfield{author}{\bibinfo{person}{Baihan Lin}.}
  \bibinfo{year}{2022}\natexlab{b}.
\newblock \showarticletitle{Regularity normalization: neuroscience-inspired
  unsupervised attention across neural network layers}.
\newblock \bibinfo{journal}{\emph{Entropy}} \bibinfo{volume}{24},
  \bibinfo{number}{1} (\bibinfo{year}{2022}), \bibinfo{pages}{59}.
\newblock


\bibitem[Lin(2022c)]%
        {sctsa}
\bibfield{author}{\bibinfo{person}{Baihan Lin}.}
  \bibinfo{year}{2022}\natexlab{c}.
\newblock \showarticletitle{Single-cell topological simplicial analysis reveals
  higher-order cellular complexity}.
\newblock \bibinfo{journal}{\emph{under review}} (\bibinfo{year}{2022}).
\newblock


\bibitem[Lin(2022d)]%
        {ttda}
\bibfield{author}{\bibinfo{person}{Baihan Lin}.}
  \bibinfo{year}{2022}\natexlab{d}.
\newblock \showarticletitle{Topological data analysis in time series: temporal
  filtration and application to single-cell genomics}.
\newblock \bibinfo{journal}{\emph{under review}} (\bibinfo{year}{2022}).
\newblock


\bibitem[Lin and Kriegeskorte(2018)]%
        {lin2018adgtic}
\bibfield{author}{\bibinfo{person}{Baihan Lin} {and} \bibinfo{person}{Nikolaus
  Kriegeskorte}.} \bibinfo{year}{2018}\natexlab{}.
\newblock \showarticletitle{{Adaptive Geo-Topological Independence Criterion}}.
\newblock \bibinfo{journal}{\emph{Preprint arXiv:1810.02923}}
  (\bibinfo{year}{2018}).
\newblock


\bibitem[Lin et~al\mbox{.}(2019)]%
        {lin2019visualizing}
\bibfield{author}{\bibinfo{person}{Baihan Lin}, \bibinfo{person}{Marieke Mur},
  \bibinfo{person}{Tim Kietzmann}, {and} \bibinfo{person}{Nikolaus
  Kriegeskorte}.} \bibinfo{year}{2019}\natexlab{}.
\newblock \showarticletitle{Visualizing representational dynamics with
  multidimensional scaling alignment}. In \bibinfo{booktitle}{\emph{Proceedings
  of Conference on Cognitive Computational Neuroscience}}.
  \bibinfo{pages}{1030--1033}.
\newblock


\bibitem[Maheswaranathan~et. al.(2019)]%
        {maheswaranathan2019universality}
\bibfield{author}{\bibinfo{person}{Niru Maheswaranathan~et. al.}}
  \bibinfo{year}{2019}\natexlab{}.
\newblock \showarticletitle{Universality and individuality in neural dynamics
  across large populations of recurrent networks}.
\newblock \bibinfo{journal}{\emph{arXiv}} (\bibinfo{year}{2019}).
\newblock


\bibitem[Mante~et. al.(2013)]%
        {mante2013context}
\bibfield{author}{\bibinfo{person}{Valerio Mante~et. al.}}
  \bibinfo{year}{2013}\natexlab{}.
\newblock \showarticletitle{{Context-dependent computation by recurrent
  dynamics in prefrontal cortex}}.
\newblock \bibinfo{journal}{\emph{Nature}} \bibinfo{volume}{503},
  \bibinfo{number}{7474} (\bibinfo{year}{2013}), \bibinfo{pages}{78}.
\newblock


\bibitem[Morcos et~al\mbox{.}(2018)]%
        {morcos2018insights}
\bibfield{author}{\bibinfo{person}{Ari Morcos}, \bibinfo{person}{Maithra
  Raghu}, {and} \bibinfo{person}{Samy Bengio}.}
  \bibinfo{year}{2018}\natexlab{}.
\newblock \showarticletitle{Insights on representational similarity in neural
  networks with canonical correlation}. In \bibinfo{booktitle}{\emph{NIPS}}.
  \bibinfo{pages}{5727--5736}.
\newblock


\bibitem[Pearson(1901)]%
        {pearson1901liii}
\bibfield{author}{\bibinfo{person}{Karl Pearson}.}
  \bibinfo{year}{1901}\natexlab{}.
\newblock \showarticletitle{{LIII. On lines and planes of closest fit to
  systems of points in space}}.
\newblock \bibinfo{journal}{\emph{Lond. Edinb.}} \bibinfo{volume}{2},
  \bibinfo{number}{11} (\bibinfo{year}{1901}), \bibinfo{pages}{559--572}.
\newblock


\bibitem[Raghu~et. al.(2017)]%
        {raghu2017svcca}
\bibfield{author}{\bibinfo{person}{Maithra Raghu~et. al.}}
  \bibinfo{year}{2017}\natexlab{}.
\newblock \showarticletitle{Svcca: Singular vector canonical correlation
  analysis for deep learning dynamics and interpretability}. In
  \bibinfo{booktitle}{\emph{NIPS}}. \bibinfo{pages}{6076--6085}.
\newblock


\bibitem[Reimann~et. al.(2017)]%
        {reimann2017cliques}
\bibfield{author}{\bibinfo{person}{Michael~W Reimann~et. al.}}
  \bibinfo{year}{2017}\natexlab{}.
\newblock \showarticletitle{Cliques of neurons bound into cavities provide a
  missing link between structure and function}.
\newblock \bibinfo{journal}{\emph{Front. Comput. Neurosci.}}
  \bibinfo{volume}{11} (\bibinfo{year}{2017}).
\newblock


\bibitem[Rigotti~et. al.(2013)]%
        {rigotti2013}
\bibfield{author}{\bibinfo{person}{Mattia Rigotti~et. al.}}
  \bibinfo{year}{2013}\natexlab{}.
\newblock \showarticletitle{{The importance of mixed selectivity in complex
  cognitive tasks}}.
\newblock \bibinfo{journal}{\emph{Nature}} \bibinfo{volume}{497},
  \bibinfo{number}{7451} (\bibinfo{year}{2013}), \bibinfo{pages}{585--590}.
\newblock
\showISSN{0028-0836}


\bibitem[Russo~et. al.(2018)]%
        {russo2018}
\bibfield{author}{\bibinfo{person}{Abigail~A Russo~et. al.}}
  \bibinfo{year}{2018}\natexlab{}.
\newblock \showarticletitle{{Motor Cortex Embeds Muscle-like Commands in an
  Untangled Population Response}}.
\newblock \bibinfo{journal}{\emph{Neuron}} \bibinfo{volume}{97},
  \bibinfo{number}{4} (\bibinfo{year}{2018}), \bibinfo{pages}{953--966.e8}.
\newblock
\showISSN{08966273}


\bibitem[Schuster and Paliwal(1997)]%
        {schuster1997bidirectional}
\bibfield{author}{\bibinfo{person}{Mike Schuster} {and}
  \bibinfo{person}{Kuldip~K Paliwal}.} \bibinfo{year}{1997}\natexlab{}.
\newblock \showarticletitle{Bidirectional recurrent neural networks}.
\newblock \bibinfo{journal}{\emph{IEEE Transactions on Signal Processing}}
  \bibinfo{volume}{45}, \bibinfo{number}{11} (\bibinfo{year}{1997}),
  \bibinfo{pages}{2673--2681}.
\newblock


\bibitem[Sussillo and Barak(2013)]%
        {sussillo2013opening}
\bibfield{author}{\bibinfo{person}{David Sussillo} {and} \bibinfo{person}{Omri
  Barak}.} \bibinfo{year}{2013}\natexlab{}.
\newblock \showarticletitle{Opening the black box: low-dimensional dynamics in
  high-dimensional recurrent neural networks}.
\newblock \bibinfo{journal}{\emph{Neural computation}} \bibinfo{volume}{25},
  \bibinfo{number}{3} (\bibinfo{year}{2013}), \bibinfo{pages}{626--649}.
\newblock


\bibitem[Williams~et. al.(2018)]%
        {Williams2018}
\bibfield{author}{\bibinfo{person}{Alex~H Williams~et. al.}}
  \bibinfo{year}{2018}\natexlab{}.
\newblock \showarticletitle{{Unsupervised discovery of demixed, low-dimensional
  neural dynamics across multiple timescales through tensor component
  analysis}}.
\newblock \bibinfo{journal}{\emph{Neuron}} \bibinfo{volume}{98},
  \bibinfo{number}{6} (\bibinfo{year}{2018}), \bibinfo{pages}{1099--1115.e8}.
\newblock
\showISSN{10974199}


\bibitem[Zomorodian and Carlsson(2005)]%
        {zomorodian2005computing}
\bibfield{author}{\bibinfo{person}{Afra Zomorodian} {and}
  \bibinfo{person}{Gunnar Carlsson}.} \bibinfo{year}{2005}\natexlab{}.
\newblock \showarticletitle{{Computing persistent homology}}.
\newblock \bibinfo{journal}{\emph{Discrete {\&} Computational Geometry}}
  \bibinfo{volume}{33}, \bibinfo{number}{2} (\bibinfo{year}{2005}),
  \bibinfo{pages}{249--274}.
\newblock


\end{thebibliography}











\end{document}